\documentclass[lettersize,journal]{IEEEtran}
\usepackage{amsmath,amsfonts}
\usepackage{array}
\usepackage{textcomp}
\usepackage{stfloats}
\usepackage{url}
\usepackage{verbatim}
\usepackage{graphicx}
\usepackage{cite}
\usepackage{bbm}
\usepackage{bbding}
\usepackage{pifont}
\usepackage{multirow}
\usepackage{booktabs}
\usepackage[normalem]{ulem} 
\usepackage{color}
\usepackage{caption}
\usepackage{subcaption}
\usepackage[ruled,linesnumbered]{algorithm2e}
\usepackage{algorithmic}
\usepackage{fancyhdr}
\renewcommand{\headrulewidth}{0pt}

\begin{document}

\title{Uncertainty-Aware Label Refinement  on Hypergraphs for Personalized Federated Facial Expression Recognition}

\author{Hu Ding, 
	Yan Yan, ~\IEEEmembership{Senior Member,~IEEE,} 
	Yang Lu, ~\IEEEmembership{Member,~IEEE,}
	Jing-Hao Xue, ~\IEEEmembership{Senior Member,~IEEE,} 
	Hanzi Wang, ~\IEEEmembership{Senior Member,~IEEE} 
\thanks{This work
was partly supported by the National Natural Science Foundation of China
under Grants 62372388, 62071404, and U21A20514, and by the
Fundamental Research Funds for the Central Universities under Grant 20720240076.} 
\thanks{Hu Ding, Yan Yan, Yang Lu, and Hanzi Wang are with the Fujian Key Laboratory of Sensing and Computing for Smart City, School of Informatics, Xiamen University, Xiamen 361102, China  (e-mail: dinghu@stu.xmu.edu.cn; yanyan@xmu.edu.cn; luyang@xmu.edu.cn;  hanzi.wang@xmu.edu.cn).
	
	Jing-Hao Xue is with the Department of Statistical Science, University College London, London, WC1E 6BT, UK (e-mail: jinghao.xue@ucl.ac.uk).
}
}

\markboth{IEEE TRANSACTIONS ON CIRCUITS AND SYSTEMS FOR VIDEO TECHNOLOGY,~Vol.~XX, No.~X, XXX~2024}
{Shell \MakeLowercase{\textit{et al.}}: A Sample Article Using IEEEtran.cls for IEEE Journals}


\maketitle

\thispagestyle{fancy}
\fancyhead{}
\lhead{}
\lfoot{}

\cfoot{Copyright~\copyright~2024 IEEE. Personal use of this material is permitted. However, permission to use this material for any other purposes must be obtained from the IEEE by sending an email to pubs-permissions@ieee.org.}

\renewcommand{\headrulewidth}{0mm}

\begin{abstract}
Most facial expression recognition (FER) models are trained on large-scale expression data with centralized learning. Unfortunately, collecting a large amount of centralized expression data is difficult in practice due to privacy concerns of facial images. In this paper, we investigate FER under the framework of personalized federated learning, which is a valuable and practical decentralized setting for real-world applications. To this end, 
we develop a novel uncertainty-\textbf{A}ware label refine\textbf{M}ent on h\textbf{Y}pergraphs (\textbf{AMY}) method. For local training, each local model consists of a backbone, an uncertainty estimation (UE) block, and an expression classification (EC) block. In the UE block, we leverage a hypergraph to model complex high-order relationships between expression samples and incorporate these relationships into uncertainty features. A personalized uncertainty estimator is then introduced to estimate reliable uncertainty weights of samples in the local client. In the EC block, we perform label propagation on the hypergraph, obtaining high-quality refined labels for retraining an expression classifier.
Based on the above, 
we effectively alleviate heterogeneous sample uncertainty across clients and learn a robust personalized FER model in each client.  Experimental results on two challenging real-world facial expression databases show that our proposed method consistently outperforms several state-of-the-art methods.
This indicates the superiority of hypergraph modeling for uncertainty estimation and label refinement on the personalized federated FER task. The source code will be released at https://github.com/mobei1006/AMY.
\end{abstract}


\section{Introduction}
\IEEEPARstart{O}{ver}  the past few decades, facial expression recognition (FER) has
received considerable attention 
with a variety of applications, including social robotics and human-computer interaction. The main goal of FER is to classify the input facial image into  one of the expression categories, including anger (AN), disgust (DI), fear (FE), happiness (HA), sadness (SA), surprise (SU), and neutrality (NE). 

A large number of FER methods \cite{cai2018island,ruan2020deep,zhang2021weakly,mo2021d3net,liu2023cross,9941139,10006835,9367174,9750079,9343871} have been developed and achieved promising performance in unconstrained scenarios. 
{These methods typically rely on large-scale  expression images to be collected and shared for centralized training. However, collecting and sharing a large number of expression images is a great challenge due to privacy concerns and the sensitivity of facial images. In many practical applications,  expression images are often distributed across local clients and are not shared}. 
Consequently, how to exploit decentralized expression data for FER merits further investigation.

\begin{figure}[t!]
	\centering
	\includegraphics[width=9cm, height=4.8cm]{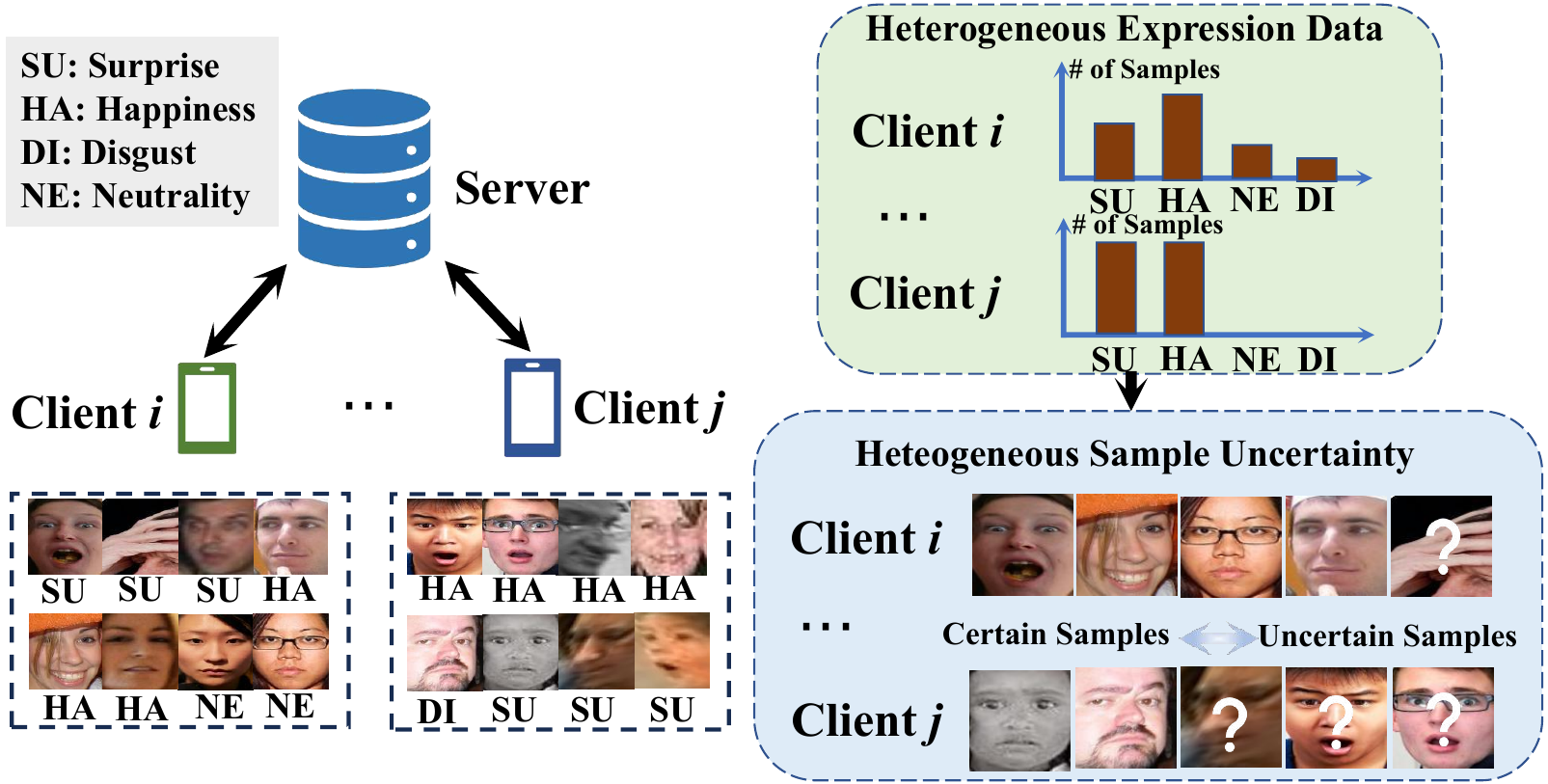}
	\caption{Illustration of the challenges of heterogeneous expression data and heterogeneous sample uncertainty. The uncertainty arises from low-quality/ambiguous expression samples and noisy labels. The degree of sample uncertainty varies for client $i$ and client $j$. The images are taken from the RAF-DB database \cite{li2017reliable}. }
	\label{fig:challenge}
\end{figure}

To address privacy concerns associated with centralized learning, federated learning (FL) \cite{mcmahan2017communication} has recently emerged as a promising decentralized learning paradigm.  FL often learns a global model by aggregating locally trained model parameters. 
Some recent methods \cite{siddiqui2022fednet,salman2022privacy,10142016,10546282} study FER within the framework of FL, capitalizing on its characteristics to learn a global FER model through local training. Regrettably, 
{due to the differences in user behavior and preferences, the data on each client may be inconsistent, making the global
model struggle to adapt to local clients effectively \cite{kulkarni2020survey}. 
Therefore, customizing personalized models tailored to individual clients becomes essential. In this paper, we study personalized federated FER (PF-FER), which aims to learn a personalized FER model in each client rather than obtaining a global model. Such a way allows each client to better adapt to local data and achieve improved recognition performance.}

For PF-FER, facial expression data scattered across different clients pose a problem of heterogeneity due to different user preferences. 
{In particular, the uncertainty arises from low-quality/ambiguous expression samples and noisy labels on local data.
This leads to heterogeneous sample uncertainty, which is detrimental to the learning of expression features in each client since the local model easily overfits some uncertain samples (such as noisy labeled samples).}
As shown in Fig.~\ref{fig:challenge}, the degree of sample uncertainty varies across different clients. 
Therefore, it is critical to suppress heterogeneous sample uncertainty in PF-FER. 

Recent FER methods \cite{wang2020suppressing,she2021dive,zhang2021relative,yang2023facial} focus on sample uncertainty suppression. One representative work is self-cure network (SCN) \cite{wang2020suppressing},  which leverages a self-attention
mechanism to learn the importance weights of images.
Based on these weights, SCN further relabels uncertain samples according to the classifier outputs. 
Lei \emph{et al.} \cite{lei2022mid} address sample uncertainty based on graph embedding. 
Note that the above FER methods work on centralized learning and are not designed for PF-FER, in which heterogeneous sample uncertainty is a main challenge. Moreover, they either ignore sample relationships or only consider the pairwise (i.e., first-order) connections  between samples. In other words, high-order relationships that involve complex interactions between multiple expression samples are not well exploited. As a result, these methods usually \textit{give unreliable uncertainty estimation and incorrect relabeling results.}  

To address the aforementioned challenges, 
we develop a novel uncertainty-\textbf{A}ware label refine\textbf{M}ent on h\textbf{Y}pergraphs (\textbf{AMY}) for 
PF-FER. In AMY, we propose to leverage hypergraph networks to model the intricate high-order relationships between multiple samples on local data, where we introduce a personalized module in each client. This enables us to effectively estimate sample uncertainty and perform label correction, achieving a robust and accurate local model. 




{Specifically, for local training, each client is composed of a backbone, an uncertainty estimation (UE) block, and an expression classification (EC) block. In the UE block, we capture the uncertainty relationships between samples based on a hypergraph network. Then, these relationships are encoded into the uncertainty features. Based on these features, a personalized uncertainty estimator is leveraged to reliably estimate the uncertainty weights of samples. We also employ a weight regularization loss to 
explicitly enlarge the differences between weights from certain samples and uncertain samples. 
In the EC block, we perform label propagation on the hypergraph, obtaining refined labels. These refined labels are combined with the model predictions to obtain final high-quality labels for retraining an expression classifier. 

After local training on each client, the local models and local class prototypes are uploaded to the server. 
The server aggregates  these models and prototypes and then sends the aggregated results back to each client for regularization, reducing the influence of heterogeneous expression data.



Our contributions can be summarized as follows: 
\begin{itemize}
	\item {To the best of our knowledge, we are the first attempt to address heterogeneous sample uncertainty in the personalized FL framework for FER (i.e., PF-FER).}
	
	\item We jointly perform uncertainty learning and label propagation based on hypergraph modeling that captures the complex relationships between expression samples. As a result, reliable uncertainty weights and high-quality label refinement results are obtained for retraining a robust FER model in the local client. 
	
	
	\item We conduct experiments on two challenging real-world facial expression databases to validate the effectiveness of our method 
	against several state-of-the-art uncertainty learning methods and personalized FL methods.
\end{itemize}

The remainder of this paper is organized as follows. First, we give the related work in Section II. Then, we describe our proposed method in detail in Section III. Next, we perform extensive experiments on the two challenging facial expression databases in Section IV. Finally, we draw the conclusion in Section V.

\section{Related Work}
In this section, we briefly review the methods closely related to our method. We first introduce facial expression recognition methods in Section II-A. Then, we introduce several state-of-the-art federated learning methods in Section II-B. Finally, we reivew some methods related to graph neural networks in Section II-C.
\subsection{Facial Expression Recognition (FER)}
With the advance of deep learning, deep neural network-based FER methods 
\cite{9367174,9941139,10006835,meng2017identity,mollahosseini2016going,ruan2021feature} 
have gained prominence.
These methods learn discriminative expression features by either designing loss functions or performing disturbance decoupling.  {Xie \emph{et al.} \cite{9367174} develop a novel triplet loss based on class-wise boundaries and multi-stage outlier suppression for FER. 
{Gu \emph{et al.}\cite{9941139} design a simple yet effective facial expression noise-tolerant  network (FENN), which explores inter-class correlations to reduce the ambiguity between  similar expression categories.}
{Chen \emph{et al.}\cite{10006835} introduce a multi-relations aware network (MRAN) that focuses on both global and local attention features, and learns multi-level relationships to obtain effective expression features.}


In recent years, some methods focus on addressing sample uncertainty in FER. 
Wang \emph{et al.} \cite{wang2020suppressing} employ a self-attention mechanism to estimate sample uncertainty. She \emph{et al.} \cite{she2021dive} 
adopt the similarity between samples and labels for uncertainty estimation. 
Zhang \emph{et al.} \cite{zhang2021relative} propose a relative uncertainty learning (RUL) method for FER. Lei \emph{et al.} \cite{lei2022mid} introduce a graph embedded uncertainty suppressing (GUS) method.  

The above methods mainly study centralized learning on large-scale expression data. Different from these methods, we investigate the PF-FER task for privacy protection. Such a task allows multiple decentralized clients to learn personalized local models collaboratively without sharing their private expression data.

\subsection{Federated Learning (FL)}
Recently, FL \cite{li2021survey,wang2021field,kairouz2021advances} has emerged as an effective decentralized learning paradigm that enables collaborative training of multiple clients in a privacy-preserving manner.
The predominant FL method is FedAvg \cite{mcmahan2017communication}, which obtains a global model by averaging model parameters trained on local clients. 
However, the performance of FedAvg is greatly affected when learning on 
non-IID data (i.e., heterogeneous data).
Numerous efforts \cite{li2020federated,luo2021no} have been made to alleviate this problem. FedProx \cite{li2020federated} rectifies model biases by incorporating a proximal term. CCVR \cite{luo2021no} retrains classifiers by sampling virtual features from an approximate Gaussian mixture model. {Zhang \emph{et al.}\cite{10142016} develop a federated spatiotemporal incremental learning method that leverages lifelong learning and federated learning to continuously optimize models on distributed edge clients.} 
{You \emph{et al.} \cite{10546282} introduce auxiliary clients involving auxiliary datasets related to federated learning tasks and generate Mixup templates for clients, addressing the privacy issues faced by Mixup-based methods.}

Some recent works study FER under the FL framework. FedNet \cite{siddiqui2022fednet} applies the federated averaging mechanism to learn a global expression classification model. FedAffect \cite{shome2021fedaffect} explores FER under the few-shot FL setting. 
{The above methods learn a global model by aggregating information from clients. However, the global model may not work well for local clients. Moreover, these methods do not account for the ubiquitous sample uncertainty on the PF-FER task, leading to sub-optimal performance.}



Instead of training a global model, personalized FL \cite{kulkarni2020survey} acknowledges the heterogeneity of data among clients by constructing a personalized model for each client. 
FedProto \cite{tan2022fedproto} aggregates the local prototypes
collected from clients, and then sends the global prototypes back to all clients to regularize local training. 
Huang \emph{et al.} \cite{huang2021personalized} introduce FedAMP, which employs federated attention message passing to enhance collaboration between similar clients. Niu and Deng  \cite{niu2022federated} introduce gradient correction for federated face recognition. Liu \emph{et al.} \cite{liu2022fedfr} learn personalized models via a decoupled feature customization module. 

Salman and Busso  \cite{salman2022privacy} are the first to study PF-FER. They aggregate local models to obtain a global model and design an unsupervised penalization strategy for video-based FER. In this paper, we also work on PF-FER, where we innovatively introduce hypergraph networks and take advantage of a personalized uncertainty estimator to mitigate the adverse effect of heterogeneous sample uncertainty in local training, aligning well with practical scenarios.

\subsection{Graph Neural Networks (GNNs)}
GNNs have shown  superiority in modeling data relationships. 
{Some FER methods \cite{ruan2021feature,chen2023multivariate} adopt 
GNNs to model pairwise (i.e., first-order) relationships between samples for classification. Nevertheless, 
such pairwise relationships are inferior in capturing complex interactions across vertices. 
Recently, hypergraph networks have been employed to model high-order correlations among data, where each hyperedge can involve multiple vertices.} Feng \emph{et al.} \cite{feng2019hypergraph} propose a hypergraph neural network (HGNN)
to encode high-order data correlations 
for representation learning. Zhang \emph{et al.} \cite{zhang2020hypergraph} introduce a hypergraph label propagation network (HLPN) 
to optimize feature embeddings.

{In our PF-FER task, each client involves the uncertainty arising from low-quality/ambiguous expression samples and noisy labels. Since the data distribution is heterogeneous across different clients,
PF-FER suffers from heterogeneous sample uncertainty. Conventional methods only consider the pairwise connections
between expression images and ignore high-order relationships (which can indicate different levels of relation). 
As a result, these methods
usually give unreliable uncertainty estimation and incorrect
relabeling results. To address this problem, 
we introduce the hypergraph neural network to model the intricate high-order relationships between multiple samples on local data. This enables us to effectively estimate sample uncertainty and perform label correction. This in turn greatly 
addresses heterogeneous sample uncertainty across local clients.}

\section{Proposed Method}
In this section, we elaborately introduce our proposed method for PF-FER. First, we give the preliminaries and notations in Section III-A. Then, we provide an overview of our method in Section III-B. Subsequently, we present technical details of the uncertainty estimation block and the expression classification block in Section III-C and Section III-D, respectively. Finally, we summarize the global training of our method in Section III-E.

\subsection{Preliminaries and Notations}
The objective of PF-FER is to collaboratively train a personalized FER model in each client by communicating between clients and the server without disclosing raw expression data. 

Suppose that we have $K$ clients and each client has $C$ expression categories, 
where the model parameters and the local data in the $k$-th client are denoted as $\mathbf{w}_k$ and $\mathcal{D}^k$, respectively. The expression data over clients are assumed to be non-IID.  
During local training, a batch of expression samples $\{\textbf{x}_i^k, y_i^k\}_{i=1}^N$, where $N$ represents the number of images in a batch, and $\textbf{x}_i^k$ and $y_i^k \in \{1,\dots, C\}$ respectively represent the $i$-th facial expression image and its corresponding label in the $k$-th client, is randomly sampled  from $\mathcal{D}^k$.    

\subsection{Overview}

Our AMY method follows two representative FL methods (a traditional FL method FedAvg \cite{mcmahan2017communication} and a personalized FL method FedProto \cite{tan2022fedproto}). It consists of a server and multiple clients. During each round of training, the server distributes its current model and global class prototypes to some selected clients. Then, each selected client trains the local model on its own expression data, and only sends the model update and its local class prototypes to the server. Note that a personalized uncertainty estimator is trained exclusively in each client and does not share its model parameters.
Next, the server aggregates model updates and local class prototypes from those selected clients. The above steps iterate several training rounds and  personalized local models are finally learned on multiple clients.  

\begin{figure*}[th!]
	\centering
	\includegraphics[scale=0.53]{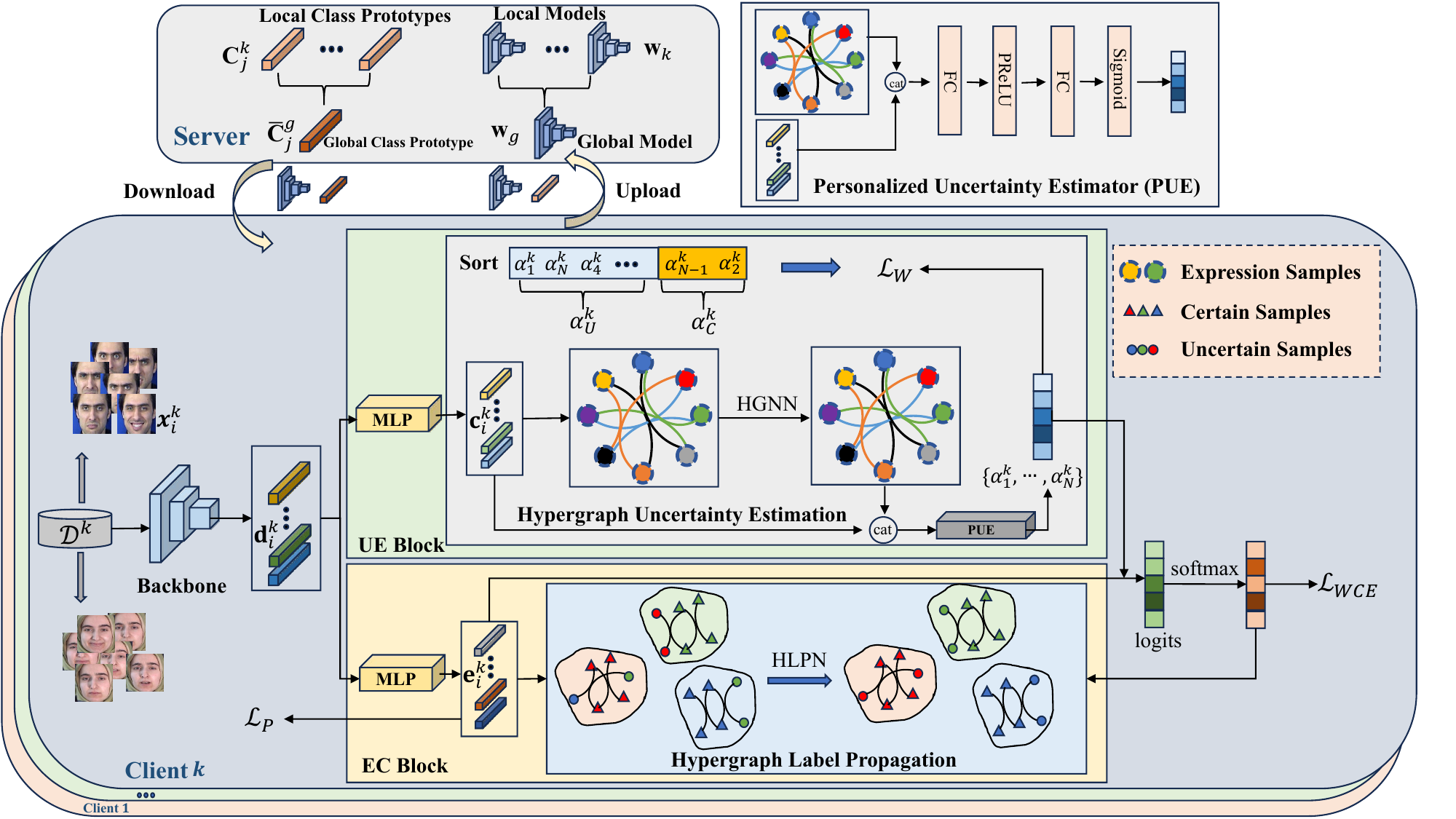}
	\caption{Overview of our proposed AMY method. Each local model is comprised of a backbone, an uncertainty estimation (UE) block, and an expression classification (EC) block. A personalized uncertainty estimator (PUE) is private to each client and not uploaded to the server, enabling personalized training. } 
	\label{fig:overview}
\end{figure*}

An overview of our proposed AMY method is shown in Fig.~\ref{fig:overview}. Each local model is comprised of three components in each client: a backbone (we adopt the simple ResNet-18 in this paper), an uncertainty estimation (UE) block,
and an expression classification (EC) block. For the local data in each client, the deep feature is initially extracted by the backbone.

{On the one hand, the UE block first employs a multi-layer perception (MLP) to extract the compact feature from the deep feature. Then, the compact feature is fed into a hypergraph neural network (HGNN) 
to model complex relationships between samples,  obtaining the relational feature. Next, the relational feature and the compact feature are concatenated to generate the uncertainty feature, which is then fed into a personalized uncertainty estimator to give an uncertainty weight for each sample. 
The personalized uncertainty estimator is private to each client and not uploaded to the server, enabling personalized training.
We also adopt a weight regularization loss \cite{wang2020suppressing} to explicitly enlarge the differences between weights from certain samples and uncertain samples.}

On the other hand, the EC block classifies the input into one expression category. It first employs an MLP to extract the expression feature from the deep feature and then 
performs label propagation on a hypergraph (HLPN), obtaining the refined labels. 
Finally, the refined labels are combined with model predictions to obtain the final label refinement results for retraining the expression classifier. 


\subsection{Uncertainty Estimation (UE) Block}

The UE block is designed to 
estimate the sample  uncertainty on the client, where higher weights will be assigned to uncertain samples (i.e., the samples with a high degree of uncertainty). 
Existing methods either employ 
a fully connected layer or 
a simple graph structure to estimate the sample uncertainty. However, such ways may not be sufficient for modeling complex high-order relationships in facial expression images, generating  unreliable uncertainty weights. To address this problem, we take advantage of hypergraph modeling, which offers great flexibility in representing data connections and exhibits superior capability in capturing complex relationships, to enable more reliable uncertainty estimation.


Given $\textbf{x}_i^k$ in the $k$-th client, we denote the deep feature extracted from the backbone as $\textbf{d}^k_i$.
Then, the compact feature is extracted by an MLP and denoted as $\textbf{c}^k_i$. The compact feature  is further fed into an $L$-layer HGNN \cite{feng2019hypergraph} for learning high-order relationships. 



{The hypergraph is denoted as $G=(V,E,W)$, where $V=\{v_1, \dots, v_N\}$ represents the vertex set (each vertex represents a compact feature corresponding to a facial expression image), $E=\{e_1, \dots, e_N\}$ represents the hyperedge set
	(each hyperedge is formed by connecting a vertex to its $K$ nearest neighbors, resulting in $N$ hyperedges connecting $K+1$ vertices each). 
    {The $K$ nearest neighbors are determined based on the Euclidean distance between vertices}. 
	$\textbf{W}_u\in \mathbb{R}^{N\times N}$ is a weight matrix denoting the weights of the hyperedges. 
	The distance between two vertices is measured by a Gaussian kernel function. 
	The hypergraph structure can be represented by an incidence matrix $\textbf{H}_u\in \mathbb{R}^{N\times N}$. For a given vertex $v\in V$ and a hyperedge $e\in E$, the element $\textbf{H}_u(v,e)$ in the incidence matrix is defined as
	\begin{equation}
		\textbf{H}_u(v,e)=\left\{
		\begin{array}{rcl}
			1 & v \in e \\
			0 & v \notin e.
		\end{array} 
		\right.
	\end{equation}
	
	For a vertex $v\in V$, its degree is defined as $d(v) = \sum_{e\in E}\textbf{W}_u(e,e)\textbf{H}_u(v,e)$. For an edge $e\in E$, its degree is defined as $\delta(e) = \sum_{v\in V}\textbf{H}_u(v,e)$. $\textbf{D}_{ue}$ and $\textbf{D}_{uv}$ denote diagonal matrices for edge and vertex degrees, respectively. 

	We consider the relational features as a hypergraph signal denoted by $\textbf{X}^l \in \mathbb{R}^{N\times C_l}$, where $C_l$ represents the dimension of the feature at the $l$-th layer. The convolution operation in the hypergraph convolutional network is formulated as
	\begin{equation}
		\textbf{X}^{l+1} = \sigma(\textbf{D}_{uv}^{-1/2}\textbf{H}_u\textbf{W}_u\textbf{D}_{ue}^{-1}\mathbf{H}_u^{\mathrm{T}} \textbf{D}_{uv}^{-1/2}\textbf{X}^l\Theta),
		\label{eq:hg}
	\end{equation}
	where $\Theta \in \mathbb{R}^{C_l\times C_{l+1}}$ denotes the learnable parameters and $\sigma$ is a non-linear activation function. $\textbf{X}^{0}= [\textbf{c}_1^k, \cdots, \textbf{c}_N^k]$ is the input signal for the hypergraph neural network.
	
	{Based on the above, we transform the compact feature $\textbf{c}^k_i$ into a relational feature $\textbf{r}^k_i$ after $L$ layers.}
	Then, the uncertainty feature $\textbf{u}^k_i$ is obtained by concatenating 
	the compact feature and the relational feature, i.e., 
	\begin{equation}
		\textbf{u}^k_i=\textrm{concat}(\textbf{c}^k_i,\textbf{r}^k_i),
	\end{equation}
	where `$\textrm{concat}(\cdot)$' denotes the concatenation operation. 

	Due to heterogeneous sample uncertainty across clients, {using} a common model to predict sample uncertainty in each client cannot guarantee the optimal performance. Instead, we make use of a personalized uncertainty estimator trained exclusively in the local client. 
	The network structure of the personalized uncertainty estimator consists of two fully connected layers, a PReLU function, and a Sigmoid function. The output of the estimator is given as 
	\begin{equation}
        {\beta^k_i=\textrm{Uncertain}(\textbf{u}^k_i),}
	\end{equation}
	where 
    {$\textrm{Uncertain}(\cdot)$ denotes the  uncertainty estimator and $\beta^k_i \in[0,1]$ denotes the uncertainty weight for the image $\textbf{x}^k_i$.}
 
  {In personalized federated learning, the personalized uncertainty estimator does not upload parameters to the server for aggregation and thus it is not affected by other clients. Instead, it only leverages local data for training, indicating that each client only relies on its unique local data to estimate  uncertainty. Such a way ensures that each client obtains personalized uncertainty weights based on the uniqueness of its data. }

	To explicitly distinguish certain samples and uncertain samples, 
	we adopt the weight regularization loss \cite{wang2020suppressing} to ensure meaningful uncertainty weights. Technically, we sort all the samples according to uncertainty weights.
 {Based on this, a threshold $\zeta$ is used to divide these samples into certain and uncertain samples.} The weight regularization loss is expressed as 
	\begin{equation}
		\mathcal{L}_W=\max\{0, \eta-(\beta_U^k-\beta_C^k)\},
	\end{equation}
	where $\eta$ is the margin and $\beta_U^k$ and $\beta_C^k$ represent the weight means of uncertain and certain samples, respectively.

{Note that each client involves the uncertainty arising from low-quality/ambiguous expression samples and noisy labels. Uncertain samples are  detrimental to the learning of expression features in each client since the local model easily overfits these samples.
{Some} existing FER methods 
use the GNN to model relationships between samples for uncertainty estimation. However, the GNN can only model pairwise relationships among samples. 
In fact, the relationships between expression images usually exhibit more complex high-order dependencies
that reflect interactions among multiple images. Hence, the HGNN is introduced to model these complex relationships, facilitating the network to identify intrinsic similarities and differences between samples, thereby more accurately estimating sample uncertainty.
}
 
	\subsection{Expression Classification (EC) Block}
	In the UE block, we estimate the uncertainty weight for each sample, where the weights for uncertain samples are higher than those for certain samples. Some uncertain samples are potentially contaminated with noisy labels. Hence, it is desirable to refine the labels of these uncertain samples, facilitating obtaining a more accurate local model. Conventional relabeling methods \cite{wang2020suppressing} perform relabeling according to model predictions. Such a strategy relies heavily on the model's inference ability. If the predicted labels by the model are not accurate enough, it can affect the model's accuracy. Ideally, we should also consider the relationships between samples to produce more accurate relabeling results. Motivated by this, we perform label propagation on the hypergraph, which involves transductive learning to update the labels according to sample relationships.

    {In the EC block, we construct another hypergraph using the {$K$-nearest neighbors algorithm} for label propagation. The expression feature $\textbf{e}^k_i$ extracted from another MLP serves as the basis for hypergraph construction. Each  expression feature is represented as a vertex, and the relationships between samples constitute hyperedges. Hence, we generate a weight matrix $\textbf{W}_e$ and an incidence matrix $\textbf{H}_e$. Accordingly,  $\textbf{D}_{ee}$ and $\textbf{D}_{ev}$ denote diagonal matrices for edge and vertex degrees, respectively.}
	
	After constructing the hypergraph, we perform label propagation on the hypergraph. Similar to  HLPN \cite{zhang2020hypergraph}, the closed-form solution for calculating the predicted scores $\hat{\textbf{F}}_u$ can be obtained by solving the following equation
	\begin{equation}
		\hat{\textbf{F}}_u=(\textbf{I}+\frac{1}{\lambda}(\textbf{I}-\textbf{D}_{ev}^{-1/2}\textbf{H}_e\textbf{W}_{e}\textbf{D}_e^{-1}\textbf{H}_e^{\textrm{T}} \textbf{D}_{ev}^{-1/2}))^{-1}\textbf{Y},
	\end{equation}
	where $\textbf{Y}$ is the original label matrix of samples, $\textbf{I}$ denotes an identity matrix, and 
	$\lambda$ is a trade-off parameter to balance the inﬂuence of the hypergraph structure regularizer.
	
	Then, we transform the score matrix $\hat{\textbf{F}}_u$ into a probability score matrix $\textbf{P}^{L,k}=[\textbf{p}^{L,k}_1,\dots,\textbf{p}_N^{L,k}]$ by the softmax function, where $\textbf{p}_i^{L,k}$ represents a vector of propagated label probabilities for the $i$-th image. 
	Meanwhile, we obtain the probability score matrix $\textbf{P}^{S,k}=[\textbf{p}^{S,k}_1,\dots,\textbf{p}_N^{S,k}]$ according to the output of the expression classifier and the softmax function, where $\textbf{p}_i^{S,k}$ represents a vector of predicted class probabilities for the $i$-th image.  We estimate the labels based on the maximum predicted class probability for each sample, expressed as 
	\begin{equation}
		l_i^{L,k}=\arg\max(\textbf{p}_i^{L,k}),
	\end{equation}
	\begin{equation}
		l_i^{S,k}=\arg\max(\textbf{p}_i^{S,k}).
	\end{equation}
	
	Meanwhile, we use uncertainty weights to select samples with unreliable labels for label refinement. The label refinement process can be defined as
	\begin{equation}
		y'=
		\begin{cases}
			l_{joint} & \text{if} \  \beta_i^k \geq \delta  \ \mathrm{and} \   l^{L,k}_i =l^{S,k}_i \\
			l_{origin} & \text{otherwise},
		\end{cases}
	\end{equation}
	where $y'$ denotes the refined label; $\delta$ is a threshold; $l_{origin}$ denotes the originally given label; {$l^{L,k}_i$ and $l^{S,k}_i$ denote the estimated labels for the $i$-th sample calculated in hypergraph label propagation and predicted by the classifier, respectively. $l_{joint}$ indicates that $l^{L,k}_i$ and $l^{S,k}_i$ are the same label.}

{Both the conventional graph-based and hypergraph-based label propagation methods 
aim to propagate labels from labeled samples to unlabeled ones based on the underlying  structure.
However, the key difference {between them lies} in the different graph structures. Conventional graph neural networks can only model pairwise relationships. In contrast, hypergraph neural networks can capture high-order relationships, where each hyperedge can connect multiple vertices. In this way, the label information can be propagated
more comprehensively across the vertices, resulting in more accurate re-labeling results (as validated in our ablation study in Section IV-B).}

    {Note that we also leverage the uncertainty weights obtained from the personalized uncertainty estimator to adjust the expression classification loss $\mathcal{L}_{WCE}$ (in the form of logit-weighted cross-entropy loss)},  which is expressed as
	\begin{equation}
		\mathcal{L}_{WCE}=-\frac{1}{N}\sum_{i=1}^N\textrm{log}\frac{e^{(1-\beta_i^k)f^k_{y_{i}^k}(\textbf{e}^k_i)}}{\sum_{j=1}^Ce^{(1-\beta_i^k)  f^k_j(\textbf{e}^k_i)}},
	\end{equation}
	where $f^k_j$ represents the $j$-th expression classifier.
	$\mathcal{L}_{WCE}$ has a positive correlation with $(1-\beta_i^k)$ \cite{wang2020suppressing}.
	
	
	Finally, the overall training loss in the $k$-th client is 
	\begin{equation}
		\mathcal{L}_{k}= \mathcal{L}_{WCE}+ \lambda_1 \mathcal{L}_W+\lambda_2 \mathcal{L}_P,
	\end{equation}
	where $\lambda_1$ and $\lambda_2$ are two balancing parameters; $\mathcal{L}_P=\frac{1}{C}\sum_{j=1}^{C} d(\textbf{C}_j^k,\bar{\textbf{C}}_j^g )$ regularizes the local training using class prototypes, in which `$d(\cdot,\cdot)$' denotes the distance measure ($L_1$ distance is used) between the local class prototype $\textbf{C}_j^k$ and the global class prototype $\bar{\textbf{C}}_j^g$ for the $j$-th expression category (see Section III-E for more details). 

	\subsection{Global Training}
	After local training, we send the local models to the server. Meanwhile, to mitigate the influence of heterogeneous data, we calculate the local class prototypes $\{\textbf{C}_j^{k}\}_{j=1}^C$ of the deep expression features for  each client and upload them to the server. The above process is given as
	\begin{equation}
		\textbf{C}_j^{k}=\frac{1}{{D}_{j}^k}\sum_{\textbf{x}_i^k\in \mathcal{D}_j^{k}}\textbf{e}^k_i,~~~~~~~j=1,\cdots, C,
	\end{equation}
	where $\mathcal{D}_{j}^k$ is a subset of the local dataset $\mathcal{D}^k$ consisting of ${D}_{j}^k$ training samples belonging to the $j$-th expression category.


	The server receives the local models and the local class prototypes, and aggregates them to obtain the global model $\textbf{w}_g$ and the global class prototypes $\{\bar{\textbf{C}}_j^g\}_{j=1}^C$, calculated as
	\begin{equation}
		\textbf{w}_g=\sum_{k\in \mathcal{S}}p_k\textbf{w}_k,
	\end{equation}
	\begin{equation}
		\bar{\textbf{C}}^g_j = \sum_{k\in \mathcal{S}}p_k\textbf{C}_{j}^k,~~~~~~~j=1,\cdots, C,
	\end{equation}
	where $p_k$ represents the weight of the $k$-th client and $\mathcal{S}$ is a randomly selected subset (i.e., active clients) from $K$ clients. 
	
	The server sends the aggregated model and the global class prototypes back to the clients. The above process is {iterated} several times, enabling each client to learn an effective personalized FER model. we summarize the overall training of our method in Algorithm~\ref{alg:algorithm}.

\begin{algorithm}[!t]
	\caption{
    {The overall training of our method}}
	\label{alg:algorithm}
	\textbf{Input}: 
    {Local datasets; the number of clients $K$; local epochs $E$; global rounds $T$.} \\
	\textbf{Output}:~
    {Personalized trained models.}\\

        {\textbf{Server}} \\
        \begin{algorithmic}[1] 
		\FOR{
        {each round $t=1, \cdots, T$}}
		\STATE 
        {Randomly select a subset of clients $S_t$ and send the global model   and global class prototypes learned at round $t-1$ to clients.}
  \\
	            \FOR{
                {each client $k \in S_t$}}
                \STATE 
                {Update the local model and local class prototypes. }
            \ENDFOR
            \STATE 
            {Obtain a new global model and global class prototypes.}
		\ENDFOR
	\end{algorithmic}

        \textbf{
        {Client}} \\
        \begin{algorithmic}[1] 
		\FOR{
        {each local epoch $i$ = $1, \cdots, E$}}
                \STATE // \textit{
                {UE Block}}
                \STATE 
                {Use the UE block to obtain the uncertainty weights of the samples.}
                \STATE 
                {Update the local model by the stochastic gradient descent (SGD) algorithm.}
                \STATE // \textit{
                {UC Block}} 
                \STATE 
                {Use the EC block to relabel uncertain samples.}
		\ENDFOR
            \STATE 
            {Calculate local class prototypes.}
            \RETURN 
            {The local model and local class prototypes.}
	\end{algorithmic}
\end{algorithm}

\begin{table*}[th!]
	\scriptsize
	\centering
        \normalsize
        \setlength{\tabcolsep}{8pt}
        \renewcommand{\arraystretch}{1.2}
 	\caption{The average recognition accuracy (\%) obtained by different variants of our method with the different values of $\alpha$ on the RAF-DB and FERPlus databases. The best results are boldfaced.}
	\begin{tabular}{l|c c c c c|c c c c c}
				\hline
				\multirow{2}{*}{Method} & \multicolumn{5}{c|}{RAF-DB} & \multicolumn{5}{c}{FERPlus} \\
				\cline{2-11}
				& $\alpha$=0.1  &  $\alpha$=0.5   & $\alpha$=1  &  $\alpha$=5 & $\alpha$=10  &  $\alpha$=0.1   &  $\alpha$=0.5   &  $\alpha$=1    & $\alpha$=5 & $\alpha$=10 \\
				\cline{1-11}
				Baseline      & 87.07              &  76.73                &  61.89   		  &     54.57       &  65.50	        & 86.36                &  69.64				   &   76.86           &  70.38			 & 73.73	\\
				
				Baseline+UE w.o.W  & 90.21             &  80.65               &  71.83  		  &     68.60      &  69.81 	        & 86.68              &  83.37 			   &   76.90            &  75.61 		 & 75.91 	\\
				
				Baseline+UE  & 89.22              &  80.52                &  73.01  		  &    65.68      &  70.90        & 87.39                &  83.36 			   &  75.80           &  76.37 			 & 76.42 	\\
				
				Baseline+UE+EC  & \textbf{90.97}     &  \textbf{80.70}       &  \textbf{73.40} 	  &     \textbf{71.97}    &  \textbf{71.52 }        & \textbf{88.23}              &  \textbf{83.93 }			   &  \textbf{77.79 }           &  \textbf{76.83} 		 & \textbf{77.30 }	\\
				
				\hline 
			\end{tabular}
	\label{tab:ablation}
\end{table*}

\section{Experimental Results}
In this section, we first introduce the experimental settings in Section IV-A. Then, we conduct ablation studies in Section IV-B. Finally, we compare our method with several state-of-the-art methods in Section IV-C.

\subsection{Experimental Settings}
We conduct experiments on two challenging real-world facial expression databases:
RAF-DB \cite{li2017reliable} and FERPlus \cite{barsoum2016training}. The RAF-DB database contains 30,000 facial images. We use seven basic expressions, including 12,271 training images and 3,068 test images. The FERPlus database, an extension of FER2013, contains 28,709 training images, 3,589 validation images, and 3,589 test images with eight expression categories.

In our experiments, all facial images are first resized to $256\times256$ and subsequently randomly cropped to $224\times224$. We adopt ResNet-18 \cite{he2016deep} as the backbone for all the competing methods. 
We use a two-layer HGNN in the UE block.
We conduct experiments using PyTorch on a single NVIDIA GeForce RTX 3090 GPU. We perform 100 communication rounds, each involving 50\% of active clients.  The database is partitioned into $K$=10 clients using a Dirichlet distribution controlled by the $\alpha$ parameter (i.e., Dir($\alpha$)), where the value of $\alpha$ is set to 0.1, 0.5, 1, 5, or 10. A lower value of $\alpha$ indicates a more heterogeneous distribution over clients.  Due to limitations of local device resources, local training is performed for only one round, with a batch size of  32. The optimization of local models uses the SGD algorithm with a learning rate of 0.10. The margin $\eta$ in Eq.~(5) is set to 0.2. 
{The threshold $\zeta$ for separating certain samples and uncertain samples is empirically set to 0.7.} 
The threshold $\delta$ in Eq.~(9) for updating labels is set to 0.6.
The values of $\lambda_1$ and $\lambda_2$ in Eq.~(11) are set to 0.8 and 1.0, respectively. 

\subsection{Ablation Studies}

Some ablation results are given in Table \ref{tab:ablation}, Fig. \ref{fig:un3}, Fig.  \ref{fig:label}, and Fig. \ref{fig:un_v}.  The baseline method is a combination of FedAvg and FedProto, where ResNet-18 and a simple fully connected layer are used for expression classification in each client. 
{Unless specified, the value of $\alpha$ is set to 5 in ablation studies.}


\noindent \textbf{Effectiveness of the Uncertainty Estimation (UE) Block.} From Table \ref{tab:ablation},
we can see that Baseline+UE achieves better performance than Baseline under the different values of $\alpha$, indicating the effectiveness of our UE block, which leverages a hypergraph neural network and a personalized uncertainty estimator to estimate reliable uncertainty weights in the local client.

\begin{figure}
	\centering
	
	\begin{subfigure}[b]{0.23\textwidth}
		\includegraphics[width=\textwidth]{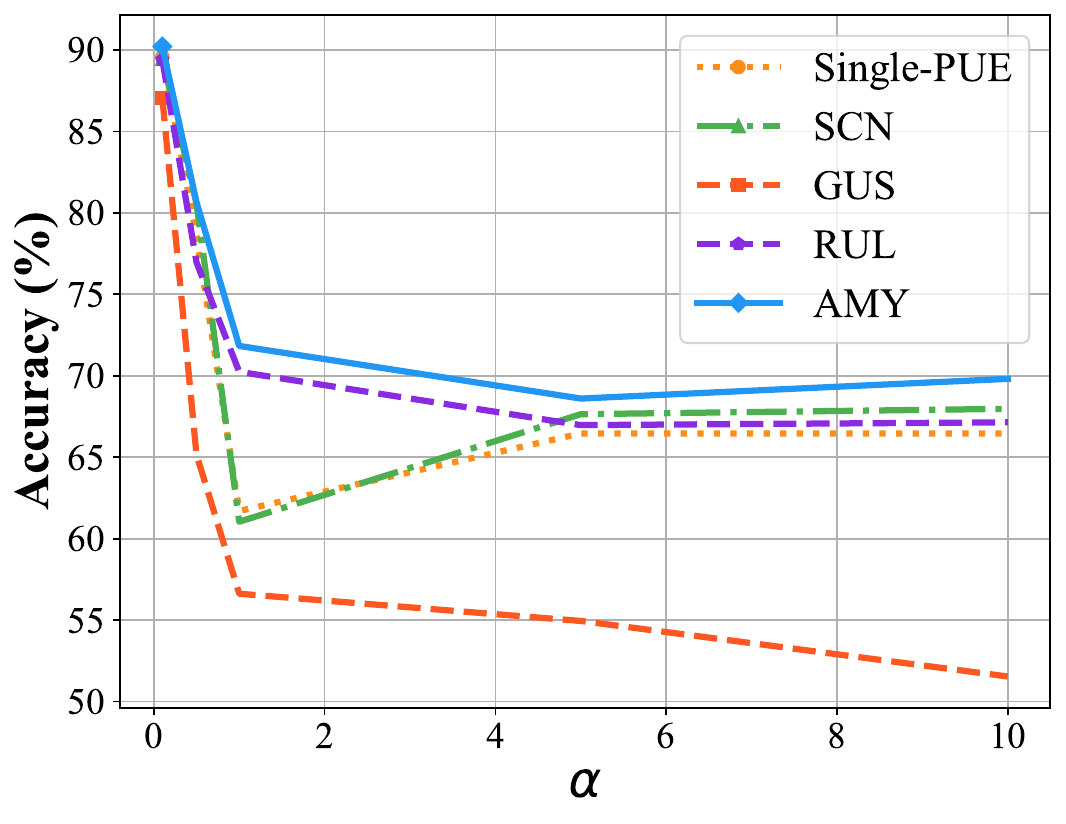}
		\caption{RAF-DB}
		\label{fig:un1}
	\end{subfigure}
	\begin{subfigure}[b]{0.23\textwidth}
		\includegraphics[width=\textwidth]{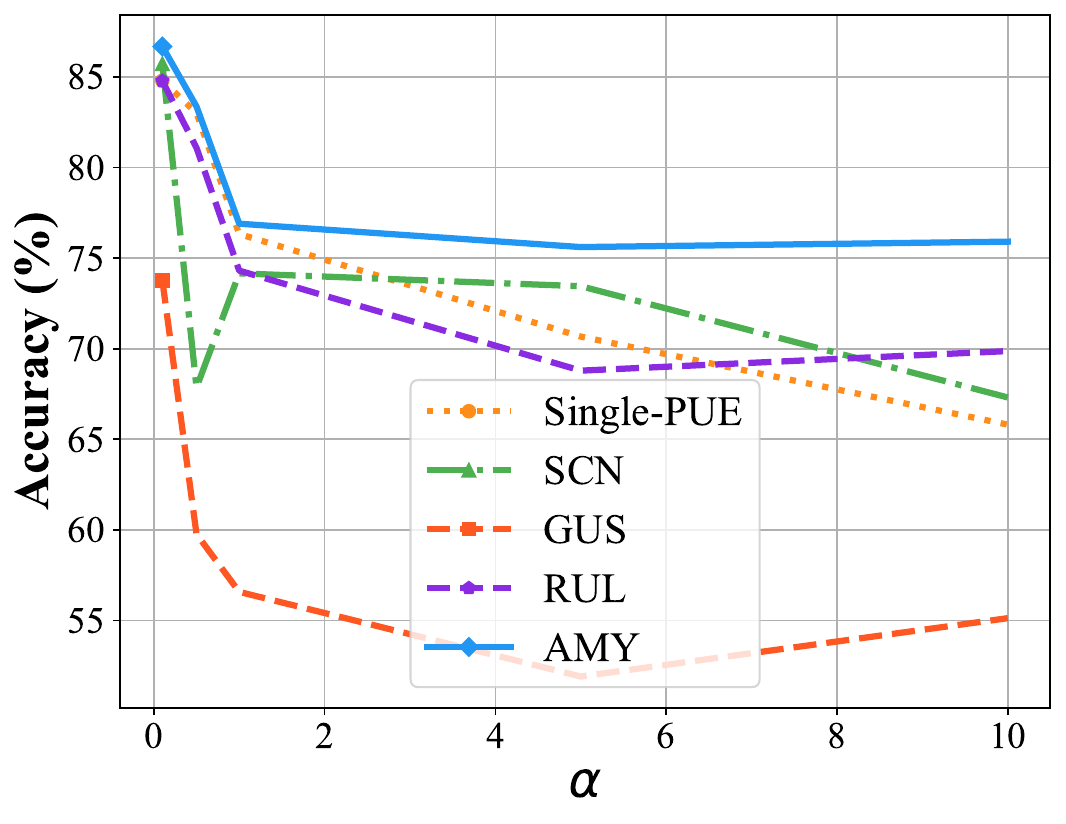}
		\caption{FERPlus}
		\label{fig:un2}
	\end{subfigure}
	
	\caption{Comparison of different competitors of the UE blo ck at the different values of $\alpha$ on the RAF-DB and FERPlus databases.}
	\label{fig:un3}
\end{figure}

\begin{figure}
	\centering
	
	\begin{subfigure}[b]{0.23\textwidth}
		\includegraphics[width=\textwidth]{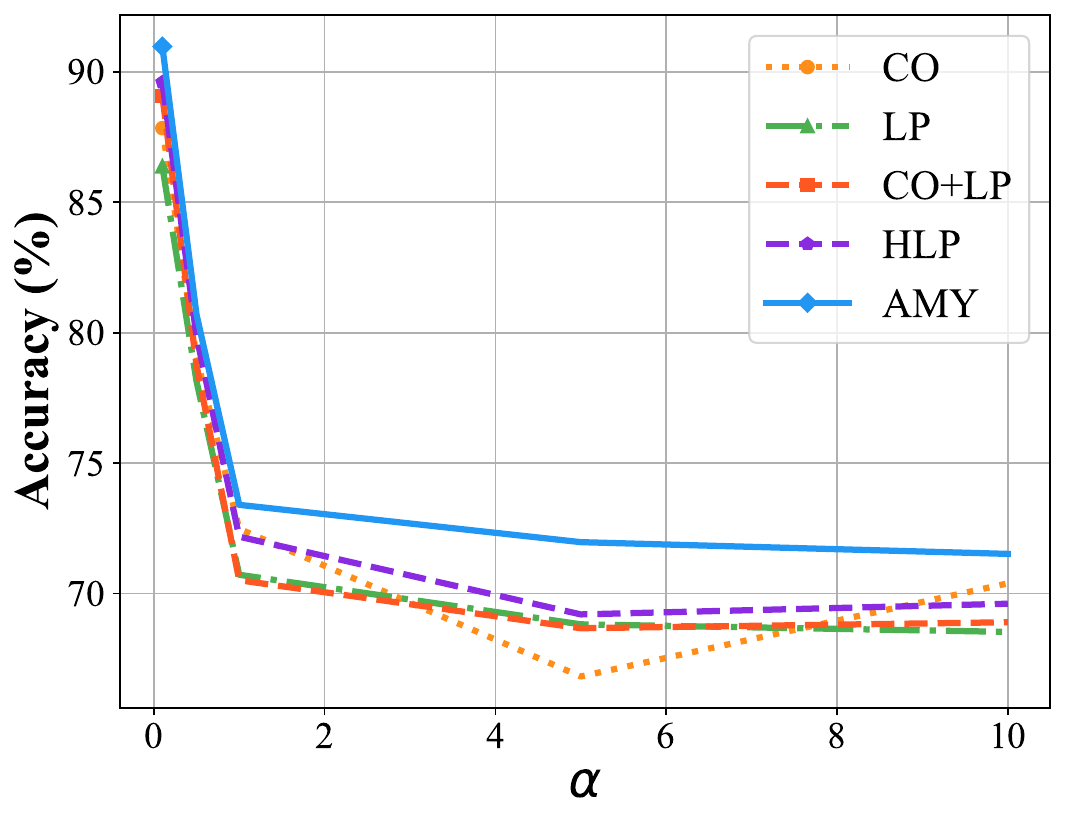}
		\caption{RAF-DB}
		\label{fig:l1}
	\end{subfigure}
	\begin{subfigure}[b]{0.23\textwidth}
		\includegraphics[width=\textwidth]{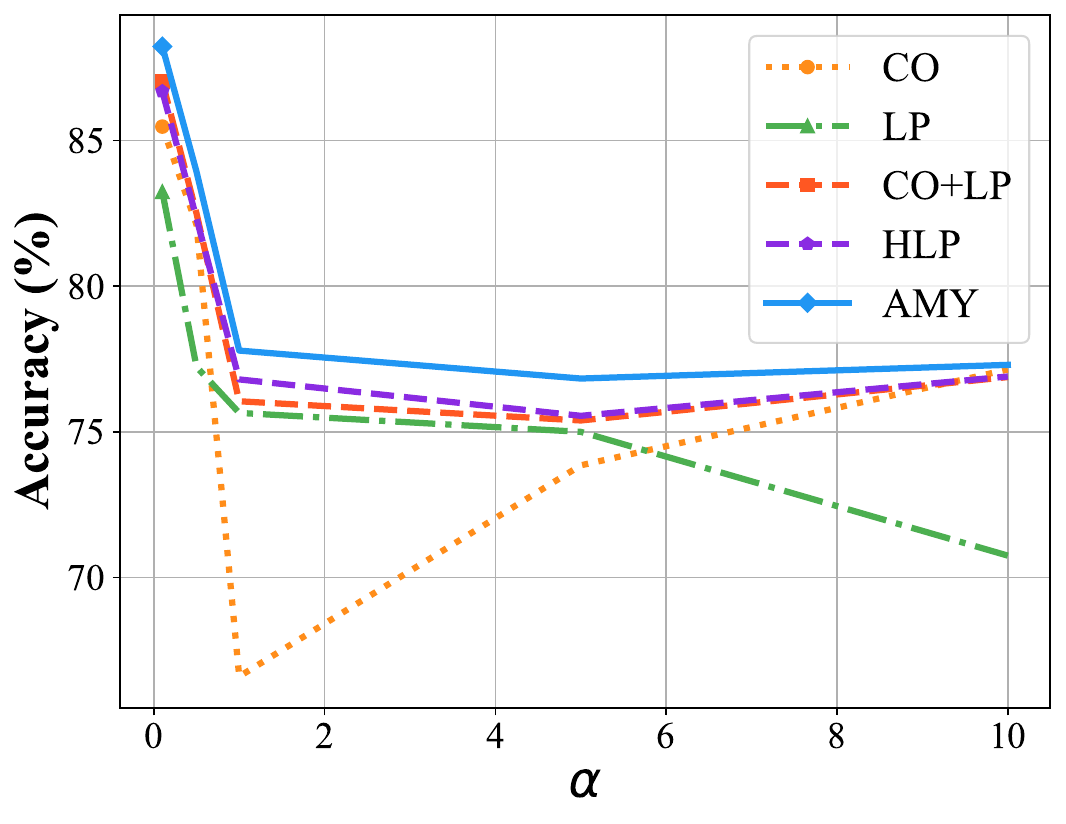}
		\caption{FERPlus}
		\label{fig:l2}
	\end{subfigure}
	
	\caption{Comparison of different competitors of the EC block at the different values of $\alpha$ on the RAF-DB and FERPlus databases.}
	\label{fig:label}
\end{figure}

To further show the advantages of high-order relationships in the hypergraph, we also evaluate some competitors (including SCN \cite{wang2020suppressing}, GUS \cite{lei2022mid}, and RUL \cite{zhang2021relative}) of the UE block, where these competitors are used to estimate the sample uncertainty. 
We also evaluate a variant of the UE block (denoted as Single-PUE) by only using the personalized uncertainty estimator without the hypergraph network. 
The comparison results are shown in Fig.~\ref{fig:un3}.
Among these competitors, our method achieves the best results under the different values of $\alpha$. Our method also outperforms Single-PUE. These results show the superiority of applying hypergraph modeling and the personalized uncertainty estimator to estimate sample uncertainty.


\noindent \textbf{Effectiveness of the Weight Regularization Loss.} 
As shown in Table \ref{tab:ablation}, without using the weight regularization loss, Baseline+UE w.o.~W gives much worse results than Baseline+UE, indicating the importance of the weight regularization loss in the UE block. By using the weight regularization loss, the model can give meaningful uncertainty weights for expression samples. These results are consistent with the experimental results on SCN \cite{wang2020suppressing}.

\begin{table*}[!t]
	\scriptsize
	\centering
        \normalsize
        \renewcommand{\arraystretch}{1.2}
 	\caption{The average recognition accuracy (\%) comparison on the RAF-DB and FERPlus databases with the different values of $\alpha$. The best results are boldfaced.}
        \begin{tabular}{l c|c c c c c|c c c c c}
				\hline
				\multirow{2}{*}{Method} & \multirow{2}{*}{Venue} & \multicolumn{5}{c|}{RAF-DB} & \multicolumn{5}{c}{FERPlus} \\
				\cline{3-12} &
				& $\alpha$=0.1  &  $\alpha$=0.5   & $\alpha$=1  &  $\alpha$=5 & $\alpha$=10  &  $\alpha$=0.1   &  $\alpha$=0.5   &  $\alpha$=1    & $\alpha$=5 & $\alpha$=10 \\
				\cline{1-12}
				Local   &--   & 88.76         & 65.53          &  51.18  		    &     42.57           & 42.75 	   			   & 83.90       		    &72.19 	  			 &   66.38  		 &  59.63			 & 54.96 	\\
				\hline
				FedAvg &AISTATS 2017 & 86.10         &  72.76          &  58.05      	  &    54.44           & 60.83 	                & 86.76                  & 66.70 			  & 75.42             &70.93  		       & 68.81 	\\
				
				FedProx &MLSys 2020 &87.16          &  73.82          &  55.35       	  &    52.92            &53.19  	            & 87.27                   & 68.62 			  & 76.85              &66.77 		       & 67.47 	\\

				\hline
				
				FedPer  & ArXiv 2019 &86.62           &  73.70           & 54.79       	  &    51.47            &60.53   	            & 87.56                   & 65.44 			  & 74.45               &65.00 	       &71.44 	\\

                    pFedMe  &NeurIPS 2020 &88.12           &  61.43            &51.70       	  &   41.46              &38.17    	            & 69.24                     & 68.48   			  &51.56            &41.43             &36.90 	\\
				
				Ditto & ICML 2021  &89.73           &  64.91           & 52.62        	  &   42.81             &44.26  	            & 84.43                    & 67.88  			  &67.16            &56.33              &57.61 	\\
				
				FedAMP  &AAAI 2021 &84.59           &  56.19             &42.10        	  &  48.71               &45.36     	       & 74.82                     & 59.62   			  &59.46             &56.37              &56.55 	\\
    
                    FedProto &AAAI 2022  &82.76            &  61.09              &50.58          	  & 42.79                &44.26      	       & 80.54                     & 67.35   			  &64.90               &58.91              &56.41 	\\
				
				FedRep &CVPR 2023  &85.57            &  78.64              &57.08         	  &  55.32               &62.61     	       & 86.47                     & 71.28  			  &76.90              &75.11              &72.16 	\\
				
				\hline
				
				Baseline &--  &87.07         &  76.73               &61.89          & 54.57             &65.50       	       & 86.36              & 69.64   			  &76.86                &70.38               &73.73 	\\
				
				Baseline+SCN &CVPR 2020  &87.99        &  80.06         &69.62     	  &68.59           &68.49              & 88.04              & 82.93  			  &76.30                &75.93              &74.68	\\
				
				Baseline+RUL & NeurIPS 2021  &89.40         &76.96         &70.24       	  &66.97             &67.14              &84.78            & 81.09 			  &74.31               &68.79              &69.87	\\

                Baseline+GUS  &ArXiv 2022 &87.04         &  65.15         &56.62      	  &54.95            &51.54              & 73.77              & 59.77 			  &56.58                &51.91              &55.13	\\
				
				AMY (Ours) & --  & \textbf{90.97}    &\textbf{80.70}    &\textbf{73.40}     &\textbf{71.97}    &\textbf{71.52 }     &\textbf{88.23} & \textbf{83.93}   &\textbf{77.79  }   &\textbf{76.83}            &\textbf{77.30 }	\\
				
				\hline 
			\end{tabular}
	\label{tab:sota}
\end{table*}

\noindent \textbf{Effectiveness of the Expression Classification (EC) Block. }
As shown in Table \ref{tab:ablation}, by employing the EC block, Baseline+UE+EC achieves better results than Baseline+UE. 
This indicates the importance of the EC block. 

To further validate the superiority of label propagation on the hypergraph, we 
evaluate several competitors of the EC block. These competitors include the traditional label propagation (LP) \cite{iscen2019label}, the simple relabeling method (CO) that performs relabeling by only using the classifier outputs as done in SCN, and the combination of the above two competitors (CO+LP).
We also evaluate a variant (HLP) of the EC block, where label refinement is only based on label propagation on the hypergraph.
The results are shown in 
Fig. \ref{fig:label}. Traditional label propagation (LP) only employs pairwise relationships between expression samples, failing to comprehensively capture complex sample relationships. The relabeling method using classifier outputs (CO)  neglects the relationships among expression data and depends solely on the model outputs for label refinement. Compared with these competitors, our method AMY consistently attains the best performance, highlighting the significance of leveraging high-order relationships between expression samples for label refinement. Note that our method achieves higher performance than HLP, showing the effectiveness of combining the label refinement results from both label propagation on the hypergraph and model predictions.

\noindent \textbf{Visualization of Uncertainty Weights.}
We visualize the uncertainty weights obtained by SCN and our method AMY during the training stage on the RAF-DB database. The results are given in Fig.~\ref{fig:un_v}. 
Our method can give more reliable uncertainty weights than SCN. 
This further validates the effectiveness of the hypergraph network and the personalized uncertainty estimator for learning uncertainty weights. 

\begin{table*}[th!]
\centering
\normalsize
\setlength{\tabcolsep}{12pt}
\renewcommand{\arraystretch}{1.2}
\caption{The classification accuracy (\%) obtained by different values of $\lambda_1$ on the RAF-DB database.}
	\begin{tabular}{c c c c c c c c c c c}
		\hline
		$\lambda_1$ & 0.1  & 0.2   & 0.3  & 0.4 & 0.5  &  0.6   & 0.7   &  0.8    & 0.9&1.0 \\
		\cline{1-11}
		acc & 70.07              &  70.98                &  71.44   		  &    70.89       &  71.02	        & 71.44                & 71.25			   &   \textbf{71.97}           &  71.88			 & 71.96	\\
		\hline 
	\end{tabular}       
\label{tab:lambda}
\end{table*}

\noindent \textbf{Influence of $\lambda_1$.} In Table \ref{tab:lambda}, we evaluate the influence of different values of  $\lambda_1$ in Eq.~(11) on the final performance. We can see that the model gives the best performance when the value of $\lambda_1$  is 0.8. This indicates the importance of weight regularization loss. 

\begin{table*}[th!]
\centering
\normalsize
\setlength{\tabcolsep}{12pt}
\renewcommand{\arraystretch}{1.2}
\caption{The classification accuracy (\%) obtained by different values of $\lambda_2$ on the RAF-DB database.}
	\begin{tabular}{c c c c c c c c c c c}
		\hline
		$\lambda_2$ & 0.1  & 0.2   & 0.3  & 0.4 & 0.5  &  0.6   & 0.7   &  0.8    & 0.9&1.0 \\
		\cline{1-11}
		acc & 69.97              &  70.76               & 70.20   		  &  68.92      &  70.81	        & 69.56              & 70.76			   &   71.21           &  71.34			 & \textbf{71.97}	\\
		\hline 
	\end{tabular}   
\label{tab:lambda2}
\end{table*}
\noindent \textbf{Influence of $\lambda_2$.} In Table \ref{tab:lambda2}, we conduct an ablation experiment on the influence of $\lambda_2$ in Eq.~(11). From Table \ref{tab:lambda2}, we can observe that when the value of $\lambda_2$ becomes larger, the performance of our method is also improved. Hence, the class prototype regularization contributes significantly to enhancing the final performance.

\begin{table}[]
\scriptsize
\centering
\normalsize
\setlength{\tabcolsep}{2pt}
\renewcommand{\arraystretch}{1.2}
\caption{The classification accuracy (\%) obtained by different values of $\eta$ on the RAF-DB database.}
	\begin{tabular}{c c c c c c c c c c }
		\hline
		$\eta$ & 0.1  & 0.2   & 0.3  & 0.4 & 0.5  &  0.6   & 0.7   &  0.8    & 0.9 \\
		\cline{1-10}
		acc & 70.96              &  \textbf{71.97}             & 69.55   		  & 60.72      &  55.67	        & 61.45              & 51.30			   &  52.40          &  53.97			\\
		\hline 
	\end{tabular}        
\label{tab:eta}
\end{table}
\noindent \textbf{Influence of $\eta$.}  In Table \ref{tab:eta}, we conduct an ablation experiment on the influence of $\eta$ in Eq. (5) (which is used to distinguish certain samples and uncertain samples). From Table \ref{tab:eta}, we can observe that the best performance is achieved when the value of $\eta$ is set to 0.2. When the value of $\eta$ is too small, it is difficult to distinguish between the two types of samples. If the value of $\eta$ is too large, the gap between the two types of samples becomes too significant, leading to incorrect uncertainty weight estimation.

\begin{table}[!t]
\scriptsize
\centering
\normalsize
\setlength{\tabcolsep}{3pt}
\renewcommand{\arraystretch}{1.2}
\caption{The classification  accuracy (\%) obtained by different values of $K$ on the RAF-DB database.}
	\begin{tabular}{c c c c c c c c c}
		\hline
		$K$ & 6  & 8   & 10  &  12 & 14  &  16  & 18   &  20 \\
		\cline{1-9}
		acc & 70.62              &  70.32               & \textbf{ 71.97  } 		  &    70.23       &  69.80	        & 70.67               &  71.14				   &   69.95\\
		\hline 
	\end{tabular}      

\label{tab:k}
\end{table}

\begin{table}[!t]
\scriptsize
\centering
\normalsize
\setlength{\tabcolsep}{2pt}
\renewcommand{\arraystretch}{1.2}
\caption{The classification accuracy (\%) obtained by different values of $\delta$ on the RAF-DB database.}
	\begin{tabular}{c c c c c c c c c c}
		\hline
		$\delta$ & 0.1  & 0.2   & 0.3  & 0.4 & 0.5  &  0.6   &0.7   & 0.8    &0.9 \\
		\cline{1-10}
		acc & 63.14             &  69.37                & 62.16    		  &   70.89       &  70.90 	        & \textbf{71.97}                &  70.56 				   &  71.64          & 71.74 	\\
		\hline 
	\end{tabular}
\label{tab:un}
\end{table}

\noindent \textbf{Influence of the Number of Neighbors ($K$) in the Hypergraph.} In Table \ref{tab:k}, we validate the influence of different neighbor numbers ($K$) in the hypergraph on the performance. We can see that a larger value of $K$ does not necessarily result in better performance, since the relationships between samples may not be well captured with too many neighbors. The optimal performance for the hypergraph is achieved when the value of $K$ is set to 10.

\noindent \textbf{Influence of the Threshold $\delta$ for Updating Labels.} In Table \ref{tab:un}, we evaluate the influence of the threshold $\delta$ in Eq.~(9) for updating labels. We can see that when the value of $\delta$ is set to 0.6, our method achieves the best results. Higher thresholds may cause the problem that 
many noisy-labeled samples are not updated. Meanwhile, lower thresholds can result in many incorrect label refinements.

\noindent 
{\textbf{Influence of the Number of Layers in the HGNN.} In the UE block, we leverage an HGNN to model high-order relationships between facial expression images.
The number of layers in the HGNN can greatly influence the performance.
We evaluate the number of layers in the HGNN on the final performance. 
The results are given in Table \ref{tab:layers}.}

{We can see that our method gives the best performance when the number of layers in the HGNN is set to 2.  Each layer of the HGNN aggregates feature information from each vertex and its neighborhoods. With the increasing number of layers, information can be aggregated from more distant vertices, allowing the model to capture broader contexts. On the one hand, when the number of layers is too small, the model fails to capture sufficient higher-order dependencies, leading to unreliable uncertainty estimation, particularly in scenarios where the clients suffer from heterogeneous sample uncertainty under personalized federated learning. On the other hand, when the number of layers is too large, the model easily suffers from overfitting since it learns information from irrelevant facial areas. Thus, the estimation of uncertainty weights is also unreliable.} 

\begin{table}[!t]
\scriptsize
\centering
\normalsize
\setlength{\tabcolsep}{10pt}
\renewcommand{\arraystretch}{1.15}
\caption{
{The classification accuracy (\%) obtained by the  different numbers of layers $\upsilon$ in the HGNN on the RAF-DB database.}}
	\begin{tabular}{c c c c c c}
		\hline
		$\upsilon$ & 1  & 2   & 3  & 4 & 5  \\
		\cline{1-6}
		acc & 70.33     & \textbf{71.97}   & 71.23   & 70.54     & 70.21 	\\
		\hline 
	\end{tabular}
\label{tab:layers}
\end{table}

\noindent 
{\textbf{Influence of the Threshold for Separating the Certain Sample Set and Uncertain Sample Set.}
In the UE block, we use a threshold to divide the whole samples into a certain sample set and an uncertain sample set. 
The certain sample set (whose uncertainty weights are low) contains high-quality expression samples that are beneficial for model training. On the contrary, the uncertain sample set (whose uncertainty weights are high) contains blurred or occluded samples, which can degrade the model performance. 
 We evaluate the influence of different thresholds $\zeta$ on the performance.  
 The results are shown in Table \ref{tab:dratio}.
Our method can obtain the best performance when the threshold is set to 0.7.}

\begin{table}[!t]
\scriptsize
\centering
\normalsize
\setlength{\tabcolsep}{2pt}
\renewcommand{\arraystretch}{1.2}
\caption{
{The classification accuracy (\%) obtained by different values of $\zeta$ on the RAF-DB database.}}
	\begin{tabular}{c c c c c c c c c c}
		\hline
		$\zeta$ & 0.1  & 0.2  & 0.3 & 0.4  &  0.5   &0.6 & 0.7   & 0.8 & 0.9 \\
		\cline{1-10}
		acc & 68.19    &  70.21    & 70.40  & 70.53  & 71.21     &  70.87    &  \textbf{71.97}   &  70.46    & 68.43  	\\
		\hline 
	\end{tabular}
\label{tab:dratio}
\end{table}

\begin{figure}[t!]
	\centering
	\includegraphics[scale=0.3]{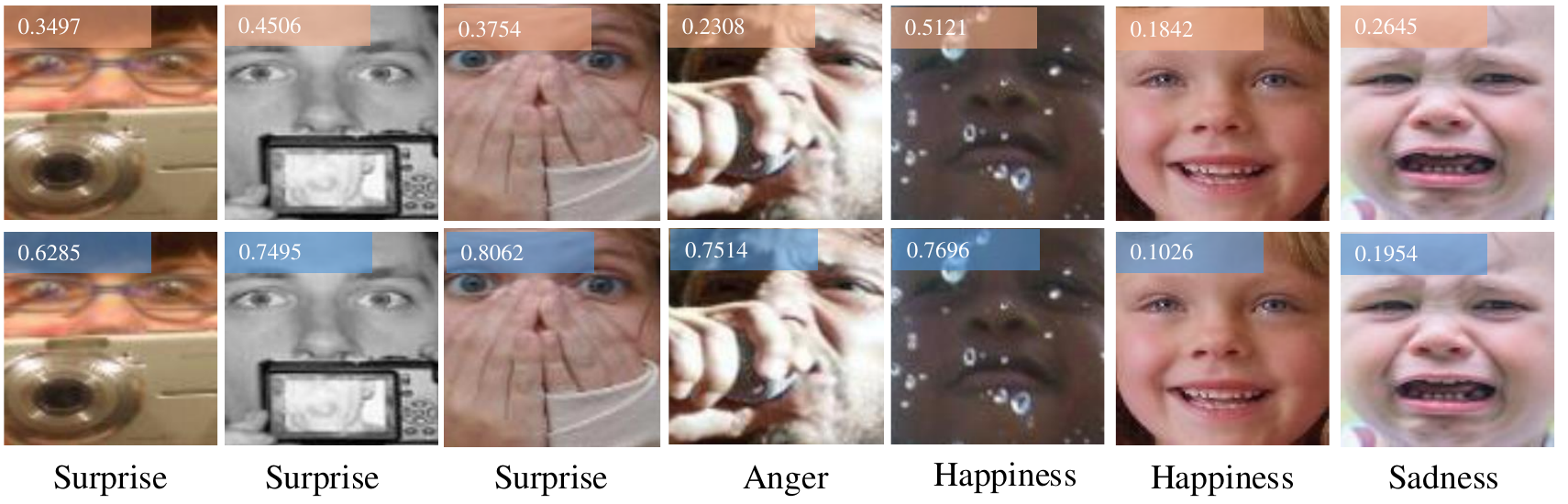}
	\caption{
    {Visualization of the uncertainty weights estimated by SCN (the first row) and our method AMY (the second row) on the RAF-DB database, where a larger weight indicates a higher degree of uncertainty for a sample.}}
	\label{fig:un_v}
\end{figure}

\noindent \textbf{Comparison Results on Centralized Learning.} In this section, we train our method on centralized learning, where only the UE block and EC block are used to train a model with all the training data. In Table \ref{tab:center}, we compare our method with SCN \cite{wang2020suppressing} and FRDL \cite{ruan2021feature}. Note that, different from federated learning, a pretrained face model \cite{wang2020suppressing} is usually used in centralized learning. Therefore, we also report the results with the pretrained model.  From Table \ref{tab:center}, we can see that our method can achieve better performance than SCN, indicating the feasibility of our method in centralized learning. Note that our method performs worse than the state-of-the-art FER method FDRL. 
{This is because our method is a very lightweight model (mainly based on ResNet-18), which is desirable in federated learning (since the memory capacity of local devices may be limited).
 On the contrary, the state-of-the-art method FDRL relies on sophisticated network design, which may not be applicable in PF-FER. The above experiments further validate 
the effectiveness of our model in the personalized federated FER task.}

\begin{table}[t!]
	\centering
        \normalsize
        \setlength{\tabcolsep}{8pt}
        \renewcommand{\arraystretch}{1.2}
 	\caption{Performance comparisons between several methods under centralized learning on the RAF-DB database. The classification accuracy (\%) is reported. }
	\begin{tabular}{l|c|c}
		\hline
		Method &RAF-DB & RAF-DB (pretrain) \\
		\hline
		SCN w.o. Relabel &  76.57 & 86.63  \\
		AMY w.o. Relabel & 77.87 & 86.95  \\
		SCN & 78.31 & 87.03 \\
		AMY &  {78.98} & {87.54} \\
		FDRL  & 80.12   & 89.47\\
		\hline
	\end{tabular}

	\label{tab:center}
\end{table}

\subsection{Comparison with State-of-the-Art Methods}
We report the performance comparison between our method and several state-of-the-art methods in Table \ref{tab:sota}.
The state-of-the-art methods include traditional FL methods (FedAvg \cite{mcmahan2017communication} and FedProx \cite{li2020federated}),
and representative personalized FL methods (FedProto \cite{tan2022fedproto}, FedPer \cite{arivazhagan2019federated}, Ditto \cite{li2021ditto}, pFedMe \cite{t2020personalized}, FedAMP \cite{huang2021personalized}, and FedRep \cite{yang2023fedrep}). In addition, we compare with personalized FL methods that incorporate  uncertainty learning, including SCN \cite{wang2020suppressing}, GUS \cite{lei2022mid}, and RUL \cite{zhang2021relative}. These methods also employ the same baseline (FedAvg+FedProto) as our method to ensure the fairness of our comparative experiments. All the competing methods are trained using publicly available codes under the same settings.
The `Local' method represents individual training for each client.

Our method AMY consistently outperforms the other competing methods. 
The `Local' method performs poorly as it only trains the models locally, lacking knowledge from other clients. Some personalized FL methods (such as FedAMP) achieve worse results than FedAvg and FedProx. This can be ascribed to the simplicity of traditional FL methods. Both traditional FL and personalized FL methods do not fully consider the challenge of heterogeneous sample uncertainty specific to the PF-FER task. 
Baseline+SCN estimates uncertainty using a single fully connected layer without considering relationships between samples. Baseline+GUS and Baseline+RUL only explore pairwise relationships between samples for uncertainty estimation, neglecting high-order relationships. On the one hand, our method addresses heterogeneous data across clients using class prototype regularization. On the other hand, our method captures complex relationships between samples using hypergraphs, which can be used for both uncertainty prediction and label refinement. The above results demonstrate that AMY is highly effective for FER in the context of personalized federated learning.

{Note that in the
traditional federated learning, a higher degree of data heterogeneity increases the difficulty of training the global model. The inconsistency in data distribution across clients can greatly influence the learning of a global model that aggregates information from local clients. However, in PF-FER, we focus on the local models. When data heterogeneity on the client is higher, the distribution of categories on that client may become more extreme (e.g., the client may only contain the images from the `happy' category). Compared with a more balanced category distribution, the local model can be more easily fine-tuned in such an extreme case, leading to improved performance in the client. Therefore, our method at $\alpha=1$ achieves better performance than our method at $\alpha=5$ on FERPlus.}

\section{
{Conclusion and Future Work} }
In this paper, we propose a novel method AMY, which learns personalized FER models over clients 
in a privacy-preserving manner, for PF-FER. AMY takes advantage of hypergraphs to model complex relationships between expression samples. Based on hypergraph modeling, 
the local model can give reliable uncertainty weights by a personalized uncertainty estimator in the UB block and generate high-quality label refinement results by label propagation in the EC block. As a result, our proposed method effectively addresses heterogeneous sample uncertainty across clients in PF-FER.  
Experiments on two challenging real-world facial expression databases validate the superiority of our method. 

{In our current method, we use a fixed threshold to divide the whole samples into a certain sample set and an uncertain sample set. However, relying on a single threshold may not effectively adapt to all clients. In addition, during the hypergraph label propagation process, we also use a fixed threshold to determine which samples need to be relabeled. In future work, we can use learnable parameters that can be adjusted based on the local data distribution and uncertainty, enabling more reasonable adaptation.}

\begin{IEEEbiography}[{\includegraphics[width=1in]{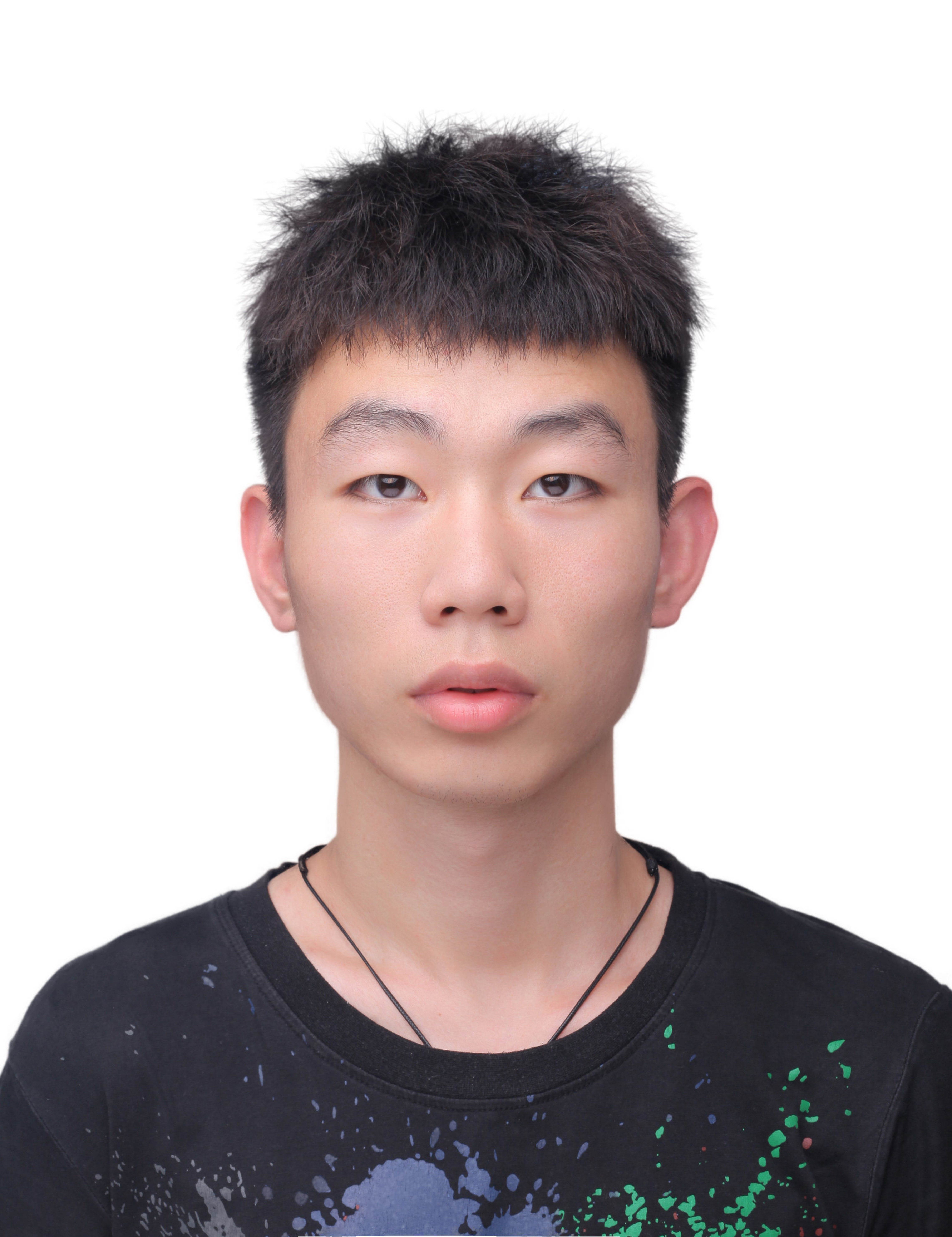}}]{Hu Ding} is currently pursuing the master's degree with the School of informatics, Xiamen University, China. His research interests include facial expression recognition and federated learning.
\end{IEEEbiography}

\begin{IEEEbiography}[{\includegraphics[width=1in]{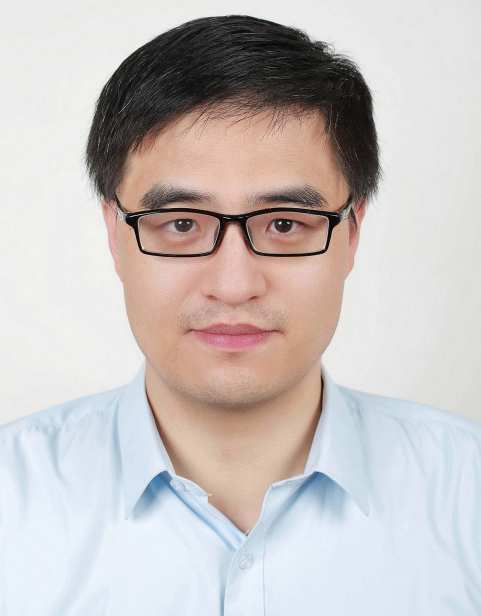}}]{Yan Yan} (Senior Member, IEEE) received the Ph.D. degree in information and communication engineering from Tsinghua University, China, in 2009. He worked as a Research Engineer with the Nokia Japan Research and Development Center from 2009 to 2010. He worked as a Project Leader with the Panasonic Singapore Laboratory in 2011. He is currently a Full Professor with the School of Informatics, Xiamen University, China. He has published around 100 papers in the international journals and conferences, including the IEEE \textsc{Transactions on Pattern Analysis and Machine Intelligence}, \textit{IJCV},  \textsc{IEEE Transactions on Image Processing}, \textsc{IEEE Transactions on Information Forensics and Security}, CVPR, ICCV, ECCV, AAAI, and ACM MM. His research interests include computer vision and pattern recognition.
\end{IEEEbiography}

\begin{IEEEbiography}[{\includegraphics[width=1in]{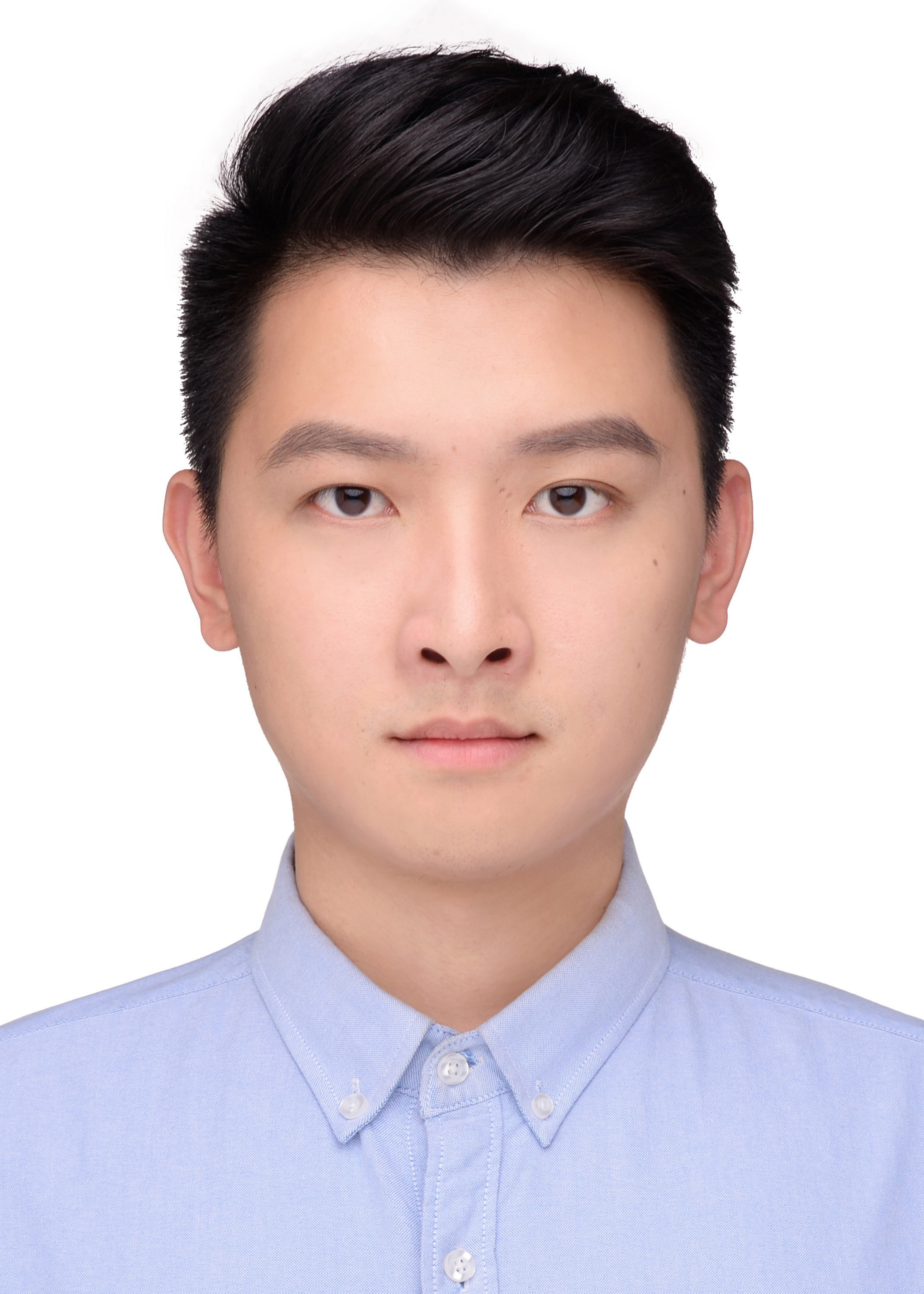}}]{Yang Lu} (Member, IEEE)  received the B.Sc. and M.Sc. degrees in software engineering from University of Macau, Macau, China, in 2012 and 2014, respectively, and the Ph.D. degree in computer science from Hong Kong Baptist University, Hong Kong, China, in 2019. He is currently an Assistant Professor with the Department of Computer Science and Technology, School of Informatics, Xiamen University, Xiamen, China. His current research interests include machine learning, deep learning, federated learning and long-tail learning.
\end{IEEEbiography}

\begin{IEEEbiography}[{\includegraphics[width=1in]{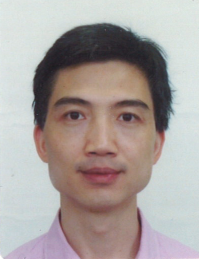}}]{Jing-Hao Xue} (Senior Member, IEEE) received the Dr.Eng. degree in signal and information processing from Tsinghua University in 1998, and the Ph.D. degree in statistics from the University of Glasgow in 2008. He is currently a Professor with the Department of Statistical Science, University College London. His research interests include statistical pattern recognition, machine learning, and computer
 vision. He received the Best Associate Editor Award of 2021 from the IEEE Transactions on Circuits and Systems for Video Technology, and the Outstanding Associate Editor Award of 2022 from the IEEE Transactions on Neural Networks and Learning Systems.
\end{IEEEbiography}

\begin{IEEEbiography}[{\includegraphics[width=1in]{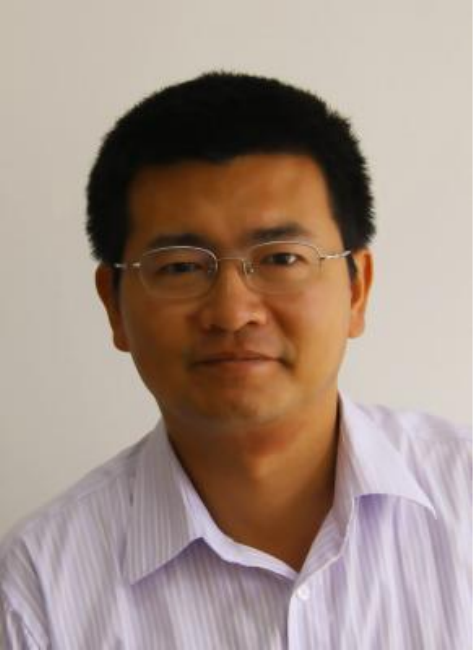}}]{Hanzi Wang} (Senior Member, IEEE) 
is currently a Distinguished Professor of ``Minjiang Scholars" in Fujian
province and a Founding Director of the Center for Pattern Analysis and Machine Intelligence at Xiamen
University, China. He received his Ph.D. degree in Computer Vision from Monash University, where he was awarded the
Douglas Lampard Electrical Engineering Research Prize and Medal for the best Ph.D. thesis. His research interests include
computer vision and pattern recognition.
\end{IEEEbiography}

\vfill

\end{document}